\documentclass[runningheads]{llncs}
\usepackage{graphicx}
\usepackage{subcaption,booktabs}

\usepackage{cite}
\usepackage{amsmath,amssymb,amsfonts}
\usepackage{algorithmic}
\usepackage{textcomp}
\usepackage{xcolor}
\usepackage{url}

\usepackage{multirow}
\usepackage{spreadtab}
\usepackage{numprint}
\usepackage{siunitx}
 \usepackage{etoolbox}

\begin{document}
\title{Multi-objective clustering: a data-driven analysis of MOCLE, MOCK and $\Delta$-MOCK}
\titlerunning{Multi-objective clustering: a data-driven analysis}
% If the paper title is too long for the running head, you can set
% an abbreviated paper title here
%
\author{Adriano Kultzak\inst{1}\orcidID{ 0000-0001-9082-9485} \and
    Cristina Y. Morimoto\inst{1}\orcidID{ 0000-0003-4122-5698} \and
	Aurora Pozo\inst{1}\orcidID{ 0000-0001-5808-3919} \and
	Marcilio C. P. de Souto\inst{2}\orcidID{ 0000-0002-7033-8328}}
\authorrunning{A. Kultzak et al.}
% First names are abbreviated in the running head.
% If there are more than two authors, 'et al.' is used.
%
\institute{Federal University of Paran\'a, Curitiba, Brazil \\
	\email{adriano\_fk@hotmail.com, cristina.morimoto@ufpr.br, aurora@inf.ufpr.br} \and
	LIFO/University of Orleans, Orleans, France\\
	\email{marcilio.desouto@univ-orleans.fr}}

\maketitle % typeset the header of the contribution
\begin{abstract}
We present a data-driven analysis of MOCK, $\Delta$-MOCK, and MOCLE. These are three closely related approaches that use multi-objective optimization for crisp clustering. More specifically, based on a collection of 12 datasets presenting different proprieties, we investigate the performance of MOCLE and MOCK compared to the recently proposed $\Delta$-MOCK. Besides performing a quantitative analysis identifying which method presents a good/poor performance with respect to another, we also conduct a more detailed analysis on why such a behavior happened. Indeed, the results of our analysis provide useful insights on the strengths and weaknesses of the methods investigated.

\end{abstract}
\section{Introduction \label{sec:initial}}

Clustering is a type of unsupervised learning whose goal is to find the underlying structure, composed of clusters, present in the data~\cite{10.5555/2535015}. Objects or observations belonging to each cluster should share some relevant property (similarity) regarding the data domain. Clustering techniques have been successfully applied in fields such as pattern analysis, decision making, data mining, image segmentation, as well as in other areas such as biology, medicine, and marketing~\cite{10.5555/2535015}.  Traditional clustering algorithms, such as $k$-means~\cite{macqueen1967some}, optimize only one clustering criterion (e.g., compactness of the clusters) and are often very effective in this purpose. However, in general, they fail for data in accordance with a different criterion (e.g., chain-like and concentric clusters in the case of $k$-means). In practice, multiple criteria, considering different aspects of the quality (e.g., compactness and connectedness) of a clustering solution (partition), frequently represent conflicting goals for an optimization method~\cite{DBLP:conf/emo/HandlK05}. This has motivated a great deal of work in which clustering is addressed as a  multi-objective problem, relying on the simultaneous optimization of multiple clustering criteria~\cite{DBLP:conf/emo/HandlK05, handl2007evolutionary, DBLP:journals/ijhis/FaceliCS07, DBLP:journals/tsmc/HruschkaCFC09, mukhopadhyay2015survey, garza2017improved, Dutta2019, DBLP:journals/kbs/ZhuXG20}. 

In particular, in this paper, by means of data-driven analysis, we revisit two popular approaches that use multi-objective optimization for crisp clustering: MOCK~\cite{DBLP:conf/emo/HandlK05, handl2007evolutionary} and MOCLE~\cite{DBLP:journals/ijhis/FaceliCS07}. We will put these two algorithms into perspective with respect to the new MOCK's version,  $\Delta$-MOCK, recently published in~\cite{garza2017improved}. The purpose is to identify the strengths and weaknesses of these algorithms so as to point out directions for future research on the development of high-quality multi-objective clustering algorithms.

The remainder of this paper is organized as follows.  Section~\ref{section:background} presents the main concepts concerning MOCK, $\Delta$-MOCK, and MOCLE, pointing out their differences, as well as discusses relevant related works. Next, Section~\ref{sec:experiment} describes the datasets considered, the specific configuration and settings of the methods compared, and the performance assessment methodology adopted. Then, in Section~\ref{sec:results}, we present and discuss the results of our experimental evaluation. Finally, Section~\ref{sec:remarks} highlights our main findings and discusses future work.

\section{Related works and Background \label{section:background}}
As this paper focuses on a data-driven investigation of MOCLE, MOCK, and $\Delta$-MOCK, it naturally shares similarities with other works~\cite{DBLP:journals/ijhis/FaceliCS07, faceli2009multi, DBLP:journals/ijon/FaceliSSC10, ANTUNES2020105971, garza2017improved}. Differently from these works, in this paper, besides performing a quantitative analysis identifying which methods present a good/poor performance concerning another, we will also conduct a more detailed/qualitative analysis to provide some insights on why such a behavior happened. Furthermore,  in order to have a better idea of the performance of $\Delta$-MOCK with respect to MOCK, as well as that of MOCLE, in this work, we performed experiments with a smaller but more diverse collection of datasets when compared to~\cite{garza2017improved, DBLP:journals/ijhis/FaceliCS07, faceli2009multi}. Finally, it is important to point out that our analysis focuses on the ability of these methods to produce a set of solutions containing high-quality partitions. Thus, with respect to MOCK and $\Delta$-MOCK, we will analyze only their clustering phase, not being concerned with the subsequent model selection phase (see Section~\ref{sec:MOCK}).

\subsection{MOCK\label{sec:MOCK}}

MOCK (Multi-Objective Clustering with automatic K-determination) is a well-known algorithm for multi-objective clustering~\cite{DBLP:conf/emo/HandlK05, handl2007evolutionary}. It is composed of two phases: clustering and model selection. The clustering phase is based on the multi-objective genetic algorithm PESA-II (Pareto envelope-based selection algorithm version II)~\cite{corne2001pesa}. To encode the solutions (partitions), MOCK uses a graph-based encoding called locus-based adjacency representation~\cite{handl2007evolutionary}: 
a solution is represented as a vector of genes,  and each gene $g_i$ can take an integer value between  $1$  and $n$. If a value $j$ is assigned to the $i$th gene, it can be interpreted as a link between the data points $i$ and $j$, i.e., $i$ and $j$ belong to the same cluster. 

The operators used are standard uniform crossover, a specialized initialization, and a neighborhood-biased mutation scheme (each object can only be linked to one of its $L$ nearest neighbors). In terms of the generation of the initial population, MOCK implements a procedure consisting of two phases: 1) By removing {\it interesting} links, partitions derived from a Minimum Spanning Tree (MST) built on the data are generated. For this,  they use a measure called {\it degree of interestingness} (DI); and 2) Generation of new partitions by removing all MST links crossing cluster boundaries of partitions initially produced by $k$-means. 

MOCK relies on the optimization of two complementary clustering criteria: overall deviation ($dev$) and connectivity ($con$), which reflect, respectively, the global concept of compactness of clusters and the more local one of the connectedness of data points~\cite{handl2007evolutionary}. More formally, the $dev$ can be defined according to \eqref{eq:dev}, where $\pi$ denotes a partition, $\mathbf{x}_i$ is an object belonging to  cluster $\mathbf{c}_k$, $\boldsymbol{\mu}_k$ is the centroid of  cluster $\mathbf{c}_k$, and $d(.,.)$ is the selected distance function.
\begin{equation}\label{eq:dev}
dev(\pi) =  \displaystyle\sum_{\mathbf{c}_k \in \pi}\displaystyle\sum_{\mathbf{x}_i \in \mathbf{c}_k}d(\mathbf{x}_i,\boldsymbol{\mu}_k),
\end{equation}

The $con$ is computed according to \eqref{eq:con}, where $n$ is the number of objects in the dataset, $L$ is the parameter that determines the number of nearest neighbors that contributes to the connectivity, $a_{ij}$ is the $j$th nearest neighbor of object $\mathbf{x}_i$, and $\mathbf{c}_k$ is a cluster that belongs to a partition $\pi$. Depending on the value chosen for parameter $L$, different partitions could present the same optimal value for $con$~\cite{garza2017improved}. As objective functions, both $dev$ and $con$ must be minimized~\cite{handl2007evolutionary}. 

\begin{equation}\label{eq:con}
    con(\pi) = \displaystyle\sum_{i=1}^{n}\displaystyle\sum_{j=1}^{L}f(\mathbf{x}_i,a_{ij}), \textrm{where } f(\mathbf{x}_i,a_{ij}) = 
    \begin{cases}
    \frac{1}{j},  \text{if}\ \nexists \mathbf{c}_k : \mathbf{x}_i, a_{ij} \in \mathbf{c}_k \\
    0,  \text{otherwise}
    \end{cases}
\end{equation}

Finally, after the clustering phase, as previously mentioned, MOCK has a model selection phase in which the algorithm provides an estimate of the quality of the partitions and determines a set of potential solutions (partitions)~\cite{handl2007evolutionary}.

\subsection{$\Delta$-MOCK \label{sec:delta-mock}}

Aiming at improving the scalability of MOCK~\cite{handl2007evolutionary}, in~\cite{garza2017improved} a new version of this algorithm, called $\Delta$-MOCK, is presented. With respect to MOCK, $\Delta$-MOCK includes changes to 1) the multi-objective optimization algorithm, 2) the objective functions, and, more fundamentally, to 3) the initialization and representation schemes. Regarding the search strategy, $\Delta$-MOCK replaces MOCK’s PESA-II~\cite{corne2001pesa} with NSGA-II (Non-dominated Sorting Genetic Algorithm II)~\cite{deb2002fast}. 

Furthermore, according to~\cite{garza2017improved}, one of the main limiting factors regarding MOCK’s scalability is the length of the genotype in the locus-based adjacency representation, which is equal to the number $n$ of objects in the dataset (see Section~\ref{sec:MOCK}). To address this problem, they introduced two alternative representations: the $\Delta$-locus and the $\Delta$-binary encodings. 
These schemes are based on the original representation of MOCK, however,   they can significantly reduce the length of the genotype by making use of information from the MST. More specifically, based on a user-defined parameter, $0 \leq \delta \leq 100$, the MST links are classified either into the set of relevant links, $\Gamma$, or into the set of non-relevant, fixed links, $\Delta$. In the new encoding proposed, only the links in $\Gamma$ must be kept, that is, a $|\Gamma|$-length genotype.  

According to~\cite{garza2017improved}, another factor that impacts MOCK’s scalability is its two-phase initialization procedure (initial population), more specifically, the second one, which is based on partitions generated by the $k$-means.  Motivated by this, they proposed an alternative procedure based on the original first phase employed by MOCK, that is, based on removing the links of the MST. 

Finally, in terms of objective functions, as MOCK, $\Delta$-MOCK uses the connectivity ($con$), but instead of MOCK’s overall deviation ($dev$), it employs the intra-cluster variance as defined in~\eqref{eq:var}, where $\pi$ denotes a partition, $n$ is the number of objects in the dataset, $\mathbf{x}_i$ is an object belonging to cluster $\mathbf{c}_k$, $\boldsymbol{\mu}_k$ is the centroid of the cluster $\mathbf{c}_k$, and $d(.,.)$ is the selected distance function. According to~\cite{garza2017improved}, although $var$ and $dev$ are conceptually similar, $var$ was chosen because it facilitates the implementation of the evaluation of a new candidate individual.

\begin{equation}\label{eq:var}
var(\pi) =  \frac{1}{n} \displaystyle\sum_{\mathbf{c}_k \in \pi}\displaystyle\sum_{\mathbf{x}_i \in \mathbf{c}_k}d(\mathbf{x}_i,\boldsymbol{\mu}_k)^{2}
\end{equation}

\subsection{MOCLE \label{sec:MOCLE}}

MOCLE (Multi-Objective Clustering Ensemble) is a clustering algorithm proposed in~\cite{DBLP:journals/ijhis/FaceliCS07} that combines characteristics from both cluster ensemble techniques and multi-objective clustering methods.  Like in cluster ensemble, starting with a diverse set of base partitions, MOCLE employs the multi-objective evolutionary algorithm to generate an approximation of the Pareto optimal set. It optimizes the same criteria as MOCK and uses a special crossover operator, which combines pairs of partition using an ensemble method. No mutation is employed. By iteratively combining pairs of partitions, MOCLE can avoid the negative influence of low-quality partitions present among the base partitions.

Finally, it is worth noting that the core ideas of MOCLE, as well as those of MOCK and $\Delta$-MOCK, are not linked to specific objective functions, crossover operator, and search algorithm. For instance, in terms of objective functions, like MOCK, MOCLE has been mainly implemented using $dev$ and $con$~\cite{DBLP:journals/ijhis/FaceliCS07, faceli2009multi, DBLP:journals/ijon/FaceliSSC10, ANTUNES2020105971}.  Concerning the crossover operators, the software available at \url{http://lasid.sor.ufscar.br/mocleproject/} implements the Meta-Clustering Algorithm (MCLA) proposed in~\cite{strehl2002cluster} and the cluster ensemble method HBGF (Hybrid Bipartite Graph Formulation)~\cite{fern2004solving}. The optimization process, like in $\Delta$-MOCK,  has been mainly performed by using NSGA-II~\cite{deb2002fast}.

\section{Experimental design \label{sec:experiment}}

\begin{table}[ht]
	\caption{Dataset characteristics: number of objects $n$, number of attributes $d$,  and number of \textit{clusters} $k^{*}$.} 
	\label{table:characteristics}
	\centering
	\begin{tabular}{ccc}\scalebox{1}{
	\begin{tabular}{llccc}
		\hline
		\textbf{Type} & \textbf{Dataset} & \textbf{$n$} & \textbf{$d$} & $k^{*}$ \\
		\hline
		\multirow{5}{*}{{\bf Artificial}} & \texttt{D31} & 3100 & 2  & 31\\
		\ & \texttt{tevc\_20\_60\_1} & 4395 & 20 & 60\\
		\ & \texttt{ds2c2sc13\_E1} & 588 & 2 &  2\\
		\ & \texttt{ds2c2sc13\_E2} & 588 & 2 &  5\\
		\ & \texttt{ds2c2sc13\_E3} & 588 & 2 &  13\\
		\hline
	\end{tabular}}
		& \hspace{5mm} &\scalebox{1}{
		\begin{tabular}{llccc}
			\hline
			\textbf{Type} & \textbf{Dataset} & \textbf{$n$} & \textbf{$d$} & $k^{*}$ \\
			\hline
		\multirow{8}{*}{{\bf Real}} & \texttt{Seeds}  & 210 & 7  & 3\\
		\ & \texttt{Golub\_E1}  & 72 & 3571 &  2\\
		\ & \texttt{Leukemia\_E2}  & 327 & 271 &  7\\
		\ & \texttt{Libras} & 360 & 90 & 15\\
		\ & \texttt{OptDigits}  & 5620 & 62 &  10\\
		\ & \texttt{Frogs\_MFCC\_E3}  & 7195 & 22 &  10\\
		\ & \texttt{UKC1} & 29463 & 2 &  11\\
		\hline  
	\end{tabular}}
\end{tabular}
\end{table}

We performed experiments with 12 datasets (available at \url{https://github.com/kultzak/multiobjective-clustering-datadriven-analysis}). Table~\ref{table:characteristics} summarizes the main characteristics of them.  {\tt D31},  {\tt tevc\_20\_60\_1} and {\tt ds2c2sc13}  are synthetic datasets specially designed to contain several different properties ~\cite{Veenman:2002:MVC:628330.628823, garza2017improved,  DBLP:journals/ijhis/FaceliCS07}. For example,  {\tt D31}  contains 31 equal size and spread clusters that are slightly overlapping and distribute randomly in a 2-dimensional space Due to the lack of spatial separation between clusters, this problem is hard to solve for algorithms based on optimizing connectedness or spatial separation. 
The dataset {\tt ds2c2sc13}  was specially designed to contain three different structures: {\tt E1}, {\tt E2} and {\tt E3}: {\tt E1} represents two well-separated clusters, which can be found  by techniques based on optimizing connectedness or compactness. In contrast, {\tt E2} and {\tt E3} combine different types of clusters that could be hard to find with techniques based only on connectedness or compactness. For example, in {\tt E2} there is a smile-shaped (non-convex) cluster, an elongated cluster, and three globular clusters. 
Concerning the real datasets, we also included datasets presenting different important characteristics. For example, {\tt Golub\_E1} and {\tt Leukemia\_E2}~\cite{faceli2009multi} have a small number of objects (distributed in clusters of very different sizes), but a large number of attributes, typical of bioinformatics data. To test the behavior of the algorithms on data with a very large number of observations, we included \texttt{UKC1}. This is one of the real datasets used in~\cite{garza2017improved} to evaluate the performance of  $\Delta$-MOCK.  \texttt{Libras}, \texttt{OptDigits} and \texttt{Frogs\_MFCC\_E3} (version with 10 classes) were obtained from the UCI Machine Learning Repository (\url{https://https://archive.ics.uci.edu/ml/datasets}).

\subsection{Performance assessment \label{sec:ARI}}
As the main indicator of clustering performance, in this work, we use the adjusted Rand index (ARI). For example, given a partition $\pi_{i}$ generated with a clustering algorithm for a dataset and a partition $\pi^{*}$ representing the underlying structure of the data (true partition/ground truth),  this measure determines the similarity between $\pi_{i}$ and $\pi^{*}$  by analyzing the pairwise co-assignment of points between the two partitions. The upper bound of ARI is 1, indicating a perfect agreement between the partitions and values near 0 or negatives corresponding to cluster agreement found by chance. Unlike the majority of other indices, ARI is not biased towards a given algorithm or number of clusters in the partition~\cite{hubert1985comparing}.

We use a non-parametric test to analyze the ARI results, the Kruskal-Wallis test with the Tukey-Kramer-Nemenyi post-hoc test~\cite{pohlert_pmcmr:_2018}  with 95\% significance. 
This test refers to the Nemenyi's non-parametric all-pairs comparison test for Kruskal-type ranked data, where the p-values are computed from the studentized range distribution of the Tukey-Kramer; this test is applied to analyze the behavior of each algorithm on a different problem (dataset).

\subsection{Experimental setup}\label{subs:exp-setup}
To execute MOCK and $\Delta$-MOCK,  we employed the same general settings as reported in~\cite{handl2007evolutionary, garza2017improved}: 1) Euclidean distance as  distance function; 2) $L=10$ nearest neighbors used during initialization, mutation, and in the definition of $con$; 3) the maximum number $k^{max}$ of clusters in the initial population was set to $2k^{*}$, where $k^{*}$ is the number of clusters in the true partition representing the dataset. Concerning the representation of the solutions for $\Delta$-MOCK, in this paper, we used the $\Delta$-locus scheme for, as pointed out in~\cite{garza2017improved}, it tends to produce better results than the  $\Delta$-binary one. Finally, regarding the length of $\Delta$-locus scheme, we set $\delta$  according to one of the heuristics employed in~\cite{garza2017improved} in which the length of the encoding is defined as a function of $\sim5\sqrt{n}$, where $n$ is the number of objects in the dataset. To execute MOCK, we used the software available at \url{https://personalpages.manchester.ac.uk/staff/Julia.Handl/mock.html}. The implementation of $\Delta$-MOCK is available at 
\url{https://github.com/garzafabre/Delta-MOCK}.

In terms of parameter configurations, for the case of MOCLE, we proceeded as follows: 1) NSGA-II as optimization method; 2) crossover operator: the cluster ensemble technique HBGF (Hybrid Bipartite Graph Formulation)~\cite{fern2004solving}; 3) objective functions: $dev$ and $con$ as in MOCK~\cite{handl2007evolutionary}; 4) Euclidean distance as  distance function; 5) 10 for the number $L$ of nearest neighbors to calculate the $con$. 
To generate the initial population, we used: $k$-means~\cite{macqueen1967some}, the Ward linkage (WL) method~\cite{ward1963hierarchical},  the shared nearest neighbor-based clustering (SNN)~\cite{ertoz2002new}, and hierarchical density based clustering (HDBSCAN)~\cite{campello2013density}. 
Also, we adjusted the parameters of such algorithms to produce partitions containing clusters in the range $\{2,2k^{*}\}$ to be consistent with MOCK/$\Delta$-MOCK's initialization.

Finally, as MOCK, $\Delta$-MOCK, and MOCLE are non-deterministic, we execute the experiments $30$ times.

\section{Results and discussion \label{sec:results}}

Table~\ref{table:ARI} presents the average and standard deviation for the ARI of best partitions, as well as for their number of clusters, found by MOCLE, MOCK, and $\Delta$-MOCK over 30 executions. In this table, for each dataset, the average of the best ARI obtained is highlighted in boldface, and the average of the numbers of clusters is indicated in parenthesis. The highlighted values also represent the best result of each dataset considering the Kruskal-Wallis/Tukey-Kramer-Nemenyi test with a two-sided p-value adjustment. 

\begin{table}[ht]
    	\caption{ARI (number of clusters) of the best partition found by MOCLE, MOCK and $\Delta$-MOCK. Average of 30 executions.} 
    \sisetup{detect-weight,mode=text}
    \renewrobustcmd{\bfseries}{\fontseries{b}\selectfont}
    \renewrobustcmd{\boldmath}{}
    \newrobustcmd{\B}{\bfseries}
	\centering 
	\sisetup{%
            table-align-uncertainty=true,
            separate-uncertainty=true,
            detect-weight=true,
            detect-inline-weight=math
        }
        \scalebox{1}{%
		\begin{tabular}{l | *{3}
		{S[table-format=1.3]@{\,\( \pm \)\,}S[table-format=1.3]
		@{\hspace{0.4em}} >{(}c @{} c<{)}}@{}
		}
			\hline
        	\textbf{Dataset}
			& \multicolumn{4}{c|}{\textbf{MOCLE}} & \multicolumn{4}{c|}{\textbf{MOCK}} &  \multicolumn{4}{c}{\textbf{$\Delta$-MOCK}} \\ \hline
			\texttt{D31} & \bfseries 0.951 & 0.003 & \hfill31.3 & & 0.843 & 0.022 & 46.7 & & 0.746 & 0.033 & 30.0 &\\
			\texttt{tevc\_20\_60\_1} & \bfseries 0.942 & 0.000 & 113.0 & & 0.794 & 0.025 & 85.2 & &  0.900 & 0.020 & 68.3  &\\
			\texttt{ds2c2sc13\_E1} & \bfseries 0.529 & 0.000 & \hfill\hfill4.0 &  & 0.381 & 0.000 & \hfill5.0 & & 0.352 & 0.000 & \hfill6.0 &\\
			\texttt{ds2c2sc13\_E2} & \bfseries 1.000 & 0.000 & \hfill\hfill5.0 & & \bfseries1.000 & 0.000 & \hfill5.0 & & 0.952 & 0.000 & \hfill6.0  &\\
			\texttt{ds2c2sc13\_E3} & \bfseries1.000 & 0.000 & \hfill13.0 & &  0.873 & 0.005 & 11.2  & & 0.872 & 0.000  & 11.0  & \\ \hline
			\texttt{Seeds} & \bfseries0.729 & 0.004 & \hfill3.8 & & 0.710 & 0.008 & \hfill3.1 & & 0.667 & 0.041 & \hfill3.8 & \\
			\texttt{Golub\_E1} & 0.473 & 0.000 & \hfill\hfill3.0 & & \bfseries0.803 & 0.096 & \hfill2.4  & & 0.541 & 0.025& 12.5 &\\
			\texttt{Leukemia\_E2} & \bfseries0.787 & 0.000 & \hfill\hfill8.0 & & 0.782 & 0.013 & \hfill9.4 & & 0.771 & 0.002 & 10.8 & \\
			\texttt{Libras} & 0.345 & 0.000 & \hfill24.0 & & \bfseries0.395 & 0.012 & 15.0  & & 0.386 & 0.005 & 14.9  &\\
			\texttt{OptDigits} & 0.820 & 0.008 & \hfill11.0  & & \bfseries0.893 & 0.013 & 21.4  & & 0.834 & 0.002 & 14.5  & \\
			\texttt{Frogs\_MFCC\_E3} & 0.816 & 0.000 & \hfill\hfill6.0  & & 0.873 & 0.022 & 15.9  & & \bfseries0.905 & 0.006 & 23.7  &\\
			\texttt{UKC1} & \bfseries1.000 & 0.000 & \hfill11.0  & & 0.998 & 0.002 & 12.2  & & 0.996 & 0.001 & 13.4  &\\ \hline
			%\textbf{AVG}
			%& \multicolumn{4}{l|}{\textbf{0.783}} & \multicolumn{4}{l|}{\textbf{0.779}} &  \multicolumn{4}{l}{\textbf{0.743}} \\ \hline
	\end{tabular}
	}
	\label{table:ARI}
\end{table}

As the initial population (base partitions) of MOCLE is generated by $k$-means, the WL method, SNN, and HDBSCAN, the results of these algorithms are summarized in Table~\ref{table:ARIBasePartitions}, in which * indicates where \texttt{Frogs\_MFCC\_E3} SNN did not produce any partition with the number of clusters in the required range.
\begin{table}[hb]
	\caption{ARI (number of clusters) of the best partition found by $k$-means, the Ward linkage method, SNN and HDBSCAN.} 
	\label{table:ARIBasePartitions}
	\centering
    \sisetup{detect-weight,mode=text}
    \renewrobustcmd{\bfseries}{\fontseries{b}\selectfont}
    \renewrobustcmd{\boldmath}{}
    \newrobustcmd{\B}{\bfseries}
	\centering 
	\sisetup{%
            table-align-uncertainty=true,
            separate-uncertainty=true,
            detect-weight=true,
            detect-inline-weight=math
        }
        \scalebox{1}{%
		\begin{tabular}{l | *{4}
		{S[table-format=1.3]
		@{\hspace{0.4em}} >{(}c @{} c<{)}}@{}
		}
			\hline
			\textbf{Dataset}
			& \multicolumn{3}{c|}{\textbf{KMeans}} & \multicolumn{3}{c|}{\textbf{Ward-link}} & \multicolumn{3}{c|}{\textbf{SNN}} & \multicolumn{3}{c}{\textbf{HDBSCAN}} \\ \hline
						\texttt{D31} & \textbf{0.953} & \hfill31 & & 0.920 &\hfill31& & 0.946 &\hfill32& & 0.806 &\hfill31& \\
			\texttt{tevc\_20\_60\_1} & 0.531 &115& & 0.586 &109& & \textbf{0.942} &113& & 0.303 &\hfill76& \\
			\texttt{ds2c2sc13\_E1} & \textbf{1.000} &\hfill\hfill2& & \textbf{1.000} &\hfill\hfill2& & 0.684 &\hfill\hfill3& & \textbf{1.000} &\hfill\hfill2& \\
			\texttt{ds2c2sc13\_E2} & 0.875 &\hfill\hfill5& & 0.854 &\hfill\hfill8& & \textbf{1.000} &\hfill\hfill5&  & \textbf{1.000} &\hfill\hfill5& \\
			\texttt{ds2c2sc13\_E3} & 0.619 &\hfill18& & 0.626 &\hfill20& & \textbf{1.000} &\hfill13&  & \textbf{1.000} &\hfill13& \\
			\hline
			\texttt{Seeds} & 0.717 &\hfill\hfill3& & \textbf{0.727} &\hfill\hfill4& & 0.691 &\hfill\hfill3& & 0.426 &\hfill\hfill3& \\
			\texttt{Golub\_E1} & \textbf{0.944} &\hfill\hfill2& & 0.689 &\hfill\hfill2& & {-0.013} &\hfill\hfill2& & 0.444 &\hfill\hfill3& \\
			\texttt{Leukemia\_E2} & 0.785 &\hfill\hfill8& & \textbf{0.787} &\hfill\hfill\hfill\hfill8& & 0.522 &\hfill\hfill6& & 0.503 &\hfill\hfill7& \\
			\texttt{Libras} & 0.322 &\hfill19& & \textbf{0.345} &\hfill24& & 0.338 &\hfill27& & 0.200 &\hfill13& \\
			\texttt{OptDigits} & 0.745 &\hfill11& & \textbf{0.814} &\hfill13& & 0.600 &\hfill16& & 0.655 &\hfill10& \\
			\texttt{Frogs\_MFCC\_E3} & 0.840 &\hfill\hfill5& & 0.816 &\hfill\hfill6&&  \hfill\hfill* & \hfill\hfill* && \textbf{0.849} &\hfill21& \\
			\texttt{UKC1} & 0.991 &\hfill11& & \textbf{1.000} &\hfill11& & 0.999 &\hfill13& & \textbf{1.000} &\hfill11& \\ \hline
	\end{tabular}}
\end{table}

\subsection{Clustering performance \label{sec:performanceARI}}

Looking at Table~\ref{table:ARI}, one can observe that, overall, across the different datasets, MOCLE, MOCK, and $\Delta$-MOCK presented distinct results concerning the ARI. Indeed, according to the Kruskal-Wallis test, there is statistical evidence that for each dataset, one algorithm could be considered better in terms of ARI than the other; where based on the Tukey-Kramer-Nemenyi post-hoc test, we observed that MOCLE performed equal or better than MOCK and $\Delta$-MOCK in all the artificial datasets and three real datasets (\texttt{Seeds}, \texttt{Leukemia\_E2} and \texttt{UKC1}); MOCK provided better results than MOCLE and $\Delta$-MOCK in three real datasets: \texttt{Golub\_E1}, \texttt{Libras} and \texttt{OptDigits}, and, it had a tied result with MOCLE regarding the artificial dataset \texttt{ds2c2sc13\_E2}; while $\Delta$-MOCK only stood out in the real dataset \texttt{Frogs\_MFCC\_E3}.  

In order to understand why some methods performed poorly for certain datasets, we will perform an analysis based on both the characteristics (bias) of the methods and the properties of the datasets. This can give us some useful insights on strengths and weaknesses, mainly concerning MOCK and $\Delta$-MOCK. 

\begin{figure}[ht]
	\centering
	\subfloat[$\Delta$-MOCK (ARI = 0.76 and  $k$ = 29)]{\includegraphics[width=0.45\textwidth]{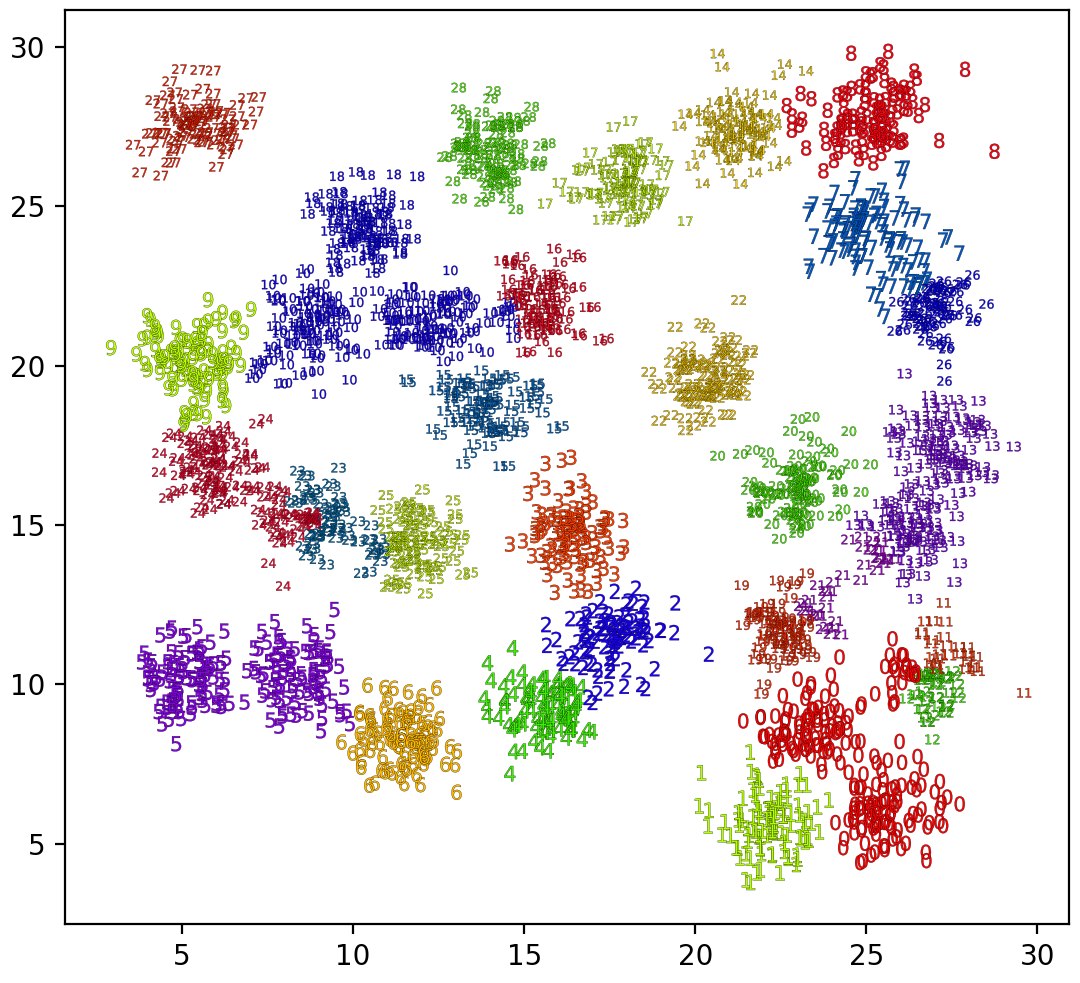}\label{fig:D31_DMOCK}}%
	\hspace{2mm}
     \subfloat[MOCK (ARI = 0.86 and  $k$ = 41)]{\includegraphics[width=0.45\textwidth]{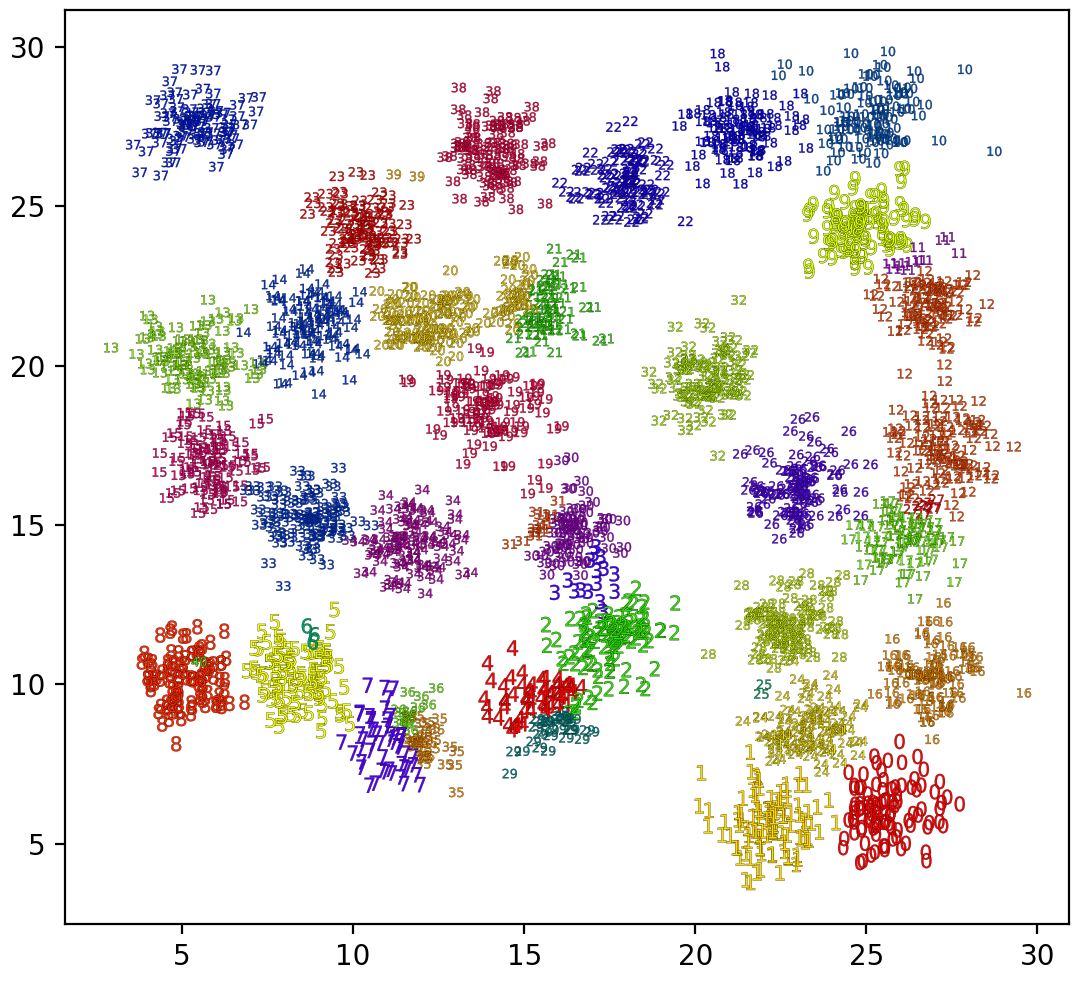}\label{fig:D31_MOCK}} 
    \caption{{\tt D31}: median best partition. Points with the same marker belong to the same cluster.}
	\label{fig:D31}
\end{figure}

For example, an analysis of the behavior of MOCK and $\Delta$-MOCK for the dataset {\tt D31}  sheds light on problems regarding the strategy to generate their initial population.  {\tt D31} contains several compact globular (Gaussian) clusters that are slightly overlapping. Thus, one would expect that any algorithm that optimizes compactness such as $k$-means and WL method should be able to yield a partition with large ARI, as can be seen in Table~\ref{table:ARIBasePartitions}. Indeed, MOCLE found, on average, high-quality partitions with ARI compatible with that returned by $k$-means (0.951 against 0.953). 
On the other hand, compared to MOCLE, on average, MOCK and $\Delta$-MOCK yielded the best solutions with much smaller ARI (respectively, 0.843 and 0.746), being the performance of $\Delta$-MOCK worse than that of MOCK. The main reason for the poor performance of $\Delta$-MOCK on {\tt D31} could be in the procedure that it employs to create the initial population. 
Its initial population is created by removing {\it interesting} links from an MST built on the data. 
In the case of {\tt D31}, as two clusters overlap, the density at the boundary of the clusters increases. As a consequence, objects in this region might have several nearest neighbors in common. This way, according to the strategy used, the links between objects in such regions would not be classified as {\it interesting} and, consequently, they would not be potential candidates to be removed and the clusters would remain merged. Indeed, a visual inspection of the median best solutions obtained with $\Delta$-MOCK for {\tt D31} shows that several clusters of the true partition were merged, forming chain-like clusters, Fig.~\ref{fig:D31_DMOCK}. MOCK also uses a strategy similar to that of $\Delta$-MOCK (see Section~\ref{sec:MOCK}), but it has solutions generated from $k$-means, added to the initial population. In this case of globular clusters, even in the presence of overlap,  the use of $k$-means within the strategy of generating the initial population of MOCK could have provided a competitive advantage over the $\Delta$-MOCK. 
For example, a visual inspection of one of the median best solutions obtained with MOCK shows that, in most cases, the clusters were not merged, Fig.~\ref{fig:D31_MOCK}. On the other hand, as can be seen in the figure, it found clusters with down to two objects (e.g., cluster 40 on the bottom of the left corner) --- this behavior will be discussed in Section~\ref{sec:size}. 

Concerning \texttt{ds2c2sc13}, independently of the multi-objective clustering algorithm, our experimental results show that their structures present some challenges in what concerns the optimization of the objective functions, mainly the $con$. Depending on the value of the $L$ used in the computation of the $con$, one can detect the two well-separated clusters in \texttt{E1} but fail to identify the small clusters present in \texttt{E3} and vice versa. Here, as in~\cite{handl2007evolutionary, garza2017improved}, we set $L$ to 10. More specifically, for \texttt{E1}, MOCLE, MOCK, and $\Delta$-MOCK could not find the two well-separated clusters presented in this structure. This happened for the true partition, $\pi^{*}$, was, in terms of the values of the objective functions, dominated by other partitions whose values for the $con$ was optimal as that of $\pi^{*}$ ($con=0$), but with smaller $dev$/$var$ (these partitions had a larger number of clusters than $\pi^{*}$). This result contrasts to that in~\cite{DBLP:journals/ijhis/FaceliCS07} in which with $L=30$ (5\% of the $n$) both MOCLE and MOCK found \texttt{E1}. For structure \texttt{E2}, like in~\cite{DBLP:journals/ijhis/FaceliCS07}, both MOCLE and MOCK found, in each of the 30 runs, a partition that corresponded to the ground truth $\pi^{*}$, that is, an ARI equals to 1. On the other hand, the best partition found by $\Delta$-MOCK in each of the runs had an ARI equals 0.952 and contained six clusters: one of the large globular clusters was split into two. In~\cite{DBLP:journals/ijhis/FaceliCS07}, with $L$ set to 30, neither MOCK nor MOCLE found \texttt{E3}. In contrast, here, in all its runs, MOCLE found a partition corresponding to the true structure in \texttt{E3} (ARI equals to 1). For the case of MOCK and $\Delta$-MOCK, the most frequent best partition in the 30 runs merged two pairs of close clusters (refinement of the globular clusters in \texttt{E2}). In summary, depending on the value of $L$ and the underlying structure in the data (e.g., a cluster with fragmented dense regions), in terms of $con$, this could mean that several different solutions will have an optimal value ($con=0$) and, as a consequence, the decision will be taken essentially based on the $dev$.

Finally, turning our attention to the real datasets, in terms of poor performance according to the ARI, there is a case that stands out:  the results of MOCLE and $\Delta$-MOCK for \texttt{Golub\_E1} (respectively, 0.473 and 0.541) when compared to that of MOCK (0.803). In the case of MOCLE, in its initial population (see Table~\ref{table:ARIBasePartitions}), there was a high-quality partition (ARI equals to 0.944) generated by $k$-means, yet such a partition was not kept in the set of final solution for it was dominated by one of the partitions generated by the SNN. For the case  $\Delta$-MOCK versus MOCK, the good performance obtained with $k$-means for \texttt{Golub\_E1} could explain why MOCK performed much better than $\Delta$-MOCK, as the MOCK initial population is built with not only MST-derived partitions like the $\Delta$-MOCK, but also with partitions generated with $k$-means.

\subsection{Number/size of clusters of the best partitions \label{sec:size}}

According to the results presented in Table~\ref{table:ARI}, although the average number of clusters of the best partitions across the different datasets, as well as the value of ARI found by the algorithms investigated, are similar, a more detailed analysis of the sizes of the clusters in the partitions can give some useful insights. To do so, for each dataset and algorithm, we chose to inspect the best median partition in terms of ARI, that is, the best partition whose ARI is the middle value among those of 30 best partitions. According to this analysis, for most datasets, MOCK and $\Delta$-MOCK generated partitions containing clusters having as few as five objects. For instance, for \texttt{Golub\_E1} the median best partition produced by $\Delta$-MOCK presented more than one cluster containing a single object. Likewise, for \texttt{OptDigits}, both algorithms yielded median best partitions with some clusters having two objects. Indeed, the problem of partitions with outlier clusters was already present in the initial population. In other words,  DI computed based on $L=10$ nearest neighbors, for most of the datasets analyzed, was not effective in avoiding the isolation of outliers.  Besides, as the $con$ is also computed based on the $L$ nearest neighbors, to a certain extent, this creates, during evolution, a bias towards solutions already present in the initial population. 

A particular case occurs in MOCLE regarding the tevc\_20\_60\_1, in which the median best partitions presented almost twice the number of clusters of the true partition. In general, this aspect could prevent us from having a good ARI result, but we observed that around 46\% of the clusters were composed of a single object, which clarifies the ARI equals 0.942. In this case, the problem of the outlier clusters also was present in the initial population, where the best partition generated by the SNN was composed of 53 clusters with a single object.

\section{Final remarks \label{sec:remarks}}

In this paper, we performed a data-driven analysis of three closely related multi-objective clustering algorithms: MOCLE, MOCK, and $\Delta$-MOCK. 
We discussed some particularities regarding the application of each algorithm across 12 datasets. 
Furthermore, in order to understand this behavior, we made an analysis based on both the characteristics (bias) of the methods and the properties of the datasets. Such an analysis gave us some useful insights on strengths and weaknesses, mainly concerning MOCK and $\Delta$-MOCK.

For example, the DI-based strategy used to generate the initial population of $\Delta$-MOCK and MOCK cannot deal efficiently with data whose clusters present some degree of overlap. Besides $\Delta$-MOCK and MOCK, this general behavior holds also for the methods in~\cite{Matake2007, Tsai2012, Zhu2018} whose initialization is also based on the MST and DI.  For the case of MOCK, at least for the case of ``Gaussian'' clusters, because of the partitions generated with $k$-means, such a problem is less pronounced.

Also, as discussed in Section~\ref{sec:size}, despite the quality in terms of ARI, for most datasets, MOCK and $\Delta$-MOCK generated partitions containing clusters having as few as five objects. Indeed, for the case of MOCK and $\Delta$-MOCK, the interaction between the use of $L$ for both generating the initial population and calculating the connectivity should be investigated further so as to provide more sound guidelines to help to choose the value of such a parameter. 

Another issue worth investigating further is the limitations of using connectivity ($con$) as objective functions. As discussed in Section~\ref{sec:MOCK} and in Section\ref{sec:performanceARI}, depending on the value chosen for  $L$, several different partitions could present the same optimal value for the connectivity ($con=0$). In a scenario like this, the connectivity would not be much discriminative, and the decision would be essentially taken based on the overall deviation. Besides MOCLE, MOCK and $\Delta$-MOCK, the previous result holds to the methods in ~\cite{Matake2007, Tsai2012} as they use $con$ and $dev$ as objective functions.

One more interesting research direction is to investigate, for example, to what extent MOCLE can benefit from the use of base partitions generated by the initialization procedure employed by $\Delta$-MOCK. An important issue not tackled in this paper concerns, after generating the set of final solutions (partitions), the model selection phase. The work in \cite{ANTUNES2020105971, DBLP:journals/kbs/ZhuXG20}, for instance, presents promising approaches to address this question.

\bibliographystyle{splncs04}
%\bibliography{references}

%\section*{Acknowledgment}

%The preferred spelling of the word ``acknowledgment'' in America is without 
%an ``e'' after the ``g''. Avoid the stilted expression ``one of us (R. B. %G.) thanks $\ldots$''. Instead, try ``R. B. G. thanks$\ldots$''. Put sponsor 
%acknowledgments in the unnumbered footnote on the first page.

%\bibliography{references}

\begin{thebibliography}{10}
	\providecommand{\url}[1]{\texttt{#1}}
	\providecommand{\urlprefix}{URL }
	\providecommand{\doi}[1]{https://doi.org/#1}
	
	\bibitem{10.5555/2535015}
	Aggarwal, C.C., Reddy, C.K.: Data Clustering: Algorithms and Applications.
	Chapman \& Hall/CRC, 1st edn. (2013)
	
	\bibitem{ANTUNES2020105971}
	Antunes, V., Sakata, T.C., Faceli, K., de~Souto, M.C.P.: Hybrid strategy for
	selecting compact set of clustering partitions. Appl. Soft Comput.
	\textbf{87},  105971 (2020). 
	
	\bibitem{campello2013density}
	Campello, R.J., Moulavi, D., Sander, J.: Density-based clustering based on
	hierarchical density estimates. In: Pacific-Asia conference on knowledge
	discovery and data mining. pp. 160--172. Springer (2013)
	
	\bibitem{corne2001pesa}
	Corne, D.W., Jerram, N.R., Knowles, J.D., Oates, M.J.: {PESA-II}: Region-based
	selection in evolutionary multiobjective optimization. In: Proc. of the
	3rd Annual Conference on Genetic and Evolutionary Computation. pp. 283--290.  (2001)
	
	\bibitem{deb2002fast}
	Deb, K., Pratap, A., Agarwal, S., Meyarivan, T.: A fast and elitist
	multiobjective genetic algorithm: {NSGA-II}. IEEE Trans. on
	evolutionary computation  \textbf{6}(2),  182--197 (2002)
	
	\bibitem{Dutta2019}
	Dutta, D., Sil, J., Dutta, P.: Automatic clustering by multi-objective genetic algorithm with numeric and categorical features. Expert Systems with 	Applications  \textbf{137} (06 2019).
	
	\bibitem{ertoz2002new}
	Ertoz, L., Steinbach, M., Kumar, V.: A new shared nearest neighbor clustering
	algorithm and its applications. In: Workshop on clustering high dimensional
	data and its applications. pp. 105--115 (2002)
	
	\bibitem{DBLP:journals/ijhis/FaceliCS07}
	Faceli, K., de~Carvalho, A.C.P., de~Souto, M.C.P.:
	Multi-objective clustering ensemble. International Journal of Hybrid
	Intelligent Systems  \textbf{4}(3),  145--156 (2007).
	
	\bibitem{DBLP:journals/ijon/FaceliSSC10}
	Faceli, K., Sakata, T.C., de~Souto, M.C.P., de~Carvalho,
	A.C.P.: Partitions selection strategy for set of clustering solutions.
	Neurocomputing  \textbf{73}(16-18),  2809--2819 (2010).
	
	\bibitem{faceli2009multi}
	Faceli, K., de~Souto, M.C.P., de~Ara{\'u}jo, D.S., de~Carvalho, A.C.:
	Multi-objective clustering ensemble for gene expression data analysis.
	Neurocomputing  \textbf{72}(13-15),  2763--2774 (2009)
	
	\bibitem{fern2004solving}
	Fern, X.Z., Brodley, C.E.: Solving cluster ensemble problems by bipartite graph
	partitioning. In: Proc. of the 21st International Conference on
	Machine learning. p.~36. ACM (2004)
	
	\bibitem{garza2017improved}
	Garza-Fabre, M., Handl, J., Knowles, J.: An improved and more scalable
	evolutionary approach to multiobjective clustering. IEEE Trans. on
	Evolutionary Computation  \textbf{22}(4),  515--535 (2017)
	
	\bibitem{handl2007evolutionary}
	Handl, J., Knowles, J.: An evolutionary approach to multiobjective clustering.
	IEEE Trans. on Evolutionary Computation  \textbf{11}(1),  56--76 (2007)
	
	\bibitem{DBLP:conf/emo/HandlK05}
	Handl, J., Knowles, J.D.: Exploiting the trade-off - the benefits of multiple
	objectives in data clustering. In: Proc. 3rd EMO International Conference. pp. 547--560 (2005). 
	
	\bibitem{DBLP:journals/tsmc/HruschkaCFC09}
	Hruschka, E.R., Campello, R.J.G.B., Freitas, A.A., de~Carvalho, A.C.P.: A survey of evolutionary algorithms for clustering. {IEEE}
	Trans. Systems, Man, and Cybernetics, Part {C}  \textbf{39}(2),  133--155
	(2009). 
	
	\bibitem{hubert1985comparing}
	Hubert, L., Arabie, P.: Comparing partitions. Journal of classification
	\textbf{2}(1),  193--218 (1985)


	\bibitem{macqueen1967some}
	MacQueen, J., et~al.: Some methods for classification and analysis of
	multivariate observations. In: Proc. of the 5th Berkeley symposium on
	mathematical statistics and probability. vol.~1, pp. 281--297.  (1967)
	
	\bibitem{Matake2007}
	Matake, N., Hiroyasu, T., Miki, M., Senda, T.: Multiobjective clustering with automatic k-determination for large-scale data. In: Proc. of the 9th Annual Conference on Genetic and Evolutionary Computation. pp. 861–868. ACM (2007). 
	
	\bibitem{mukhopadhyay2015survey}
	Mukhopadhyay, A., Maulik, U., Bandyopadhyay, S.: A survey of multiobjective
	evolutionary clustering. ACM Computing Surveys (CSUR)  \textbf{47}(4), ~61
	(2015)
	
	\bibitem{pohlert_pmcmr:_2018}
	Pohlert, T., and Pohlert, M. T.:  Package ‘PMCMR’. R package version, 1(0) (2018)
	
	
	\bibitem{strehl2002cluster}
	Strehl, A., Ghosh, J.: Cluster ensembles---a knowledge reuse framework for
	combining multiple partitions. Journal of machine learning research
	\textbf{3}(Dec),  583--617 (2002)
	
	\bibitem{Tsai2012}
	Tsai, C.,  Chen, W., Chiang, M.: A modified multiobjective EA-based clustering algorithm with automatic determination of the number of clusters, In: 2012 IEEE International Conference on Systems, Man, and Cybernetics, pp. 2833-2838 (2012)
	
	\bibitem{Veenman:2002:MVC:628330.628823}
	Veenman, C.J., Reinders, M.J.T., Backer, E., Backer, E.: A maximum variance 	cluster algorithm. IEEE Trans. Pattern Anal. Mach. Intell.  \textbf{24}(9), 	1273--1280 (Sep 2002). 
	
	\bibitem{ward1963hierarchical}
	Ward~Jr, J.H.: Hierarchical grouping to optimize an objective function. Journal
	of the American statistical association  \textbf{58}(301),  236--244 (1963)
	
	\bibitem{DBLP:journals/kbs/ZhuXG20}
	Zhu, S., Xu, L., Goodman, E.D.: Evolutionary multi-objective automatic
	clustering enhanced with quality metrics and ensemble strategy. Knowledge
	Based Systems  \textbf{188} (2020). 
	
	\bibitem{Zhu2018}
	Zhu, S., Xu, L., Cao, L.: A study of automatic clustering based on evolutionary many-objective optimization. In: Proc. of the Genetic and Evolutionary Computation Conference, pp. 173-–174. ACM. (2018)
	
\end{thebibliography}

\end{document}